\documentclass[letterpaper]{article} 
\usepackage{aaai2027}  
\usepackage[hyphens]{url}  
\usepackage{graphicx} 
\usepackage{natbib}  
\usepackage{caption} 
\usepackage{algorithm}
\usepackage{algorithmic}

\usepackage{newfloat}
\usepackage{listings}
\DeclareCaptionStyle{ruled}{labelfont=normalfont,labelsep=colon,strut=off} 
\floatstyle{ruled}
\newfloat{listing}{tb}{lst}{}
\floatname{listing}{Listing}

\usepackage{booktabs}
\usepackage{amsmath}
\usepackage{amssymb}
\usepackage{bm}
\usepackage{amsthm}
\usepackage{multirow}
\usepackage{xcolor}
\usepackage{adjustbox}

\newtheorem{proposition}{Proposition}

\title{CoReFlow: Consistency-Relaxed Flow Maps for Probabilistic Traffic Forecasting}
\author{
    Pengnan Chi\textsuperscript{\rm 1}\corresponding,
    Zhiguo Zhang\textsuperscript{\rm 1},
    Xiaoliang Ma\textsuperscript{\rm 1}
}
\affiliations{
    \textsuperscript{\rm 1} KTH Royal Institute of Technology, Stockholm, Sweden\\
    pengnan@kth.se, zhiguo@kth.se, liang@kth.se
}

\begin{document}

\maketitle

\begin{abstract}
Probabilistic traffic forecasting is essential for reliable Intelligent Transportation Systems (ITS), as it captures uncertainties in complex road networks. 
Recent diffusion- and flow-based generative models have facilitated uncertainty modeling through a multi-step sampling process. 
However, beyond slow inference, this process is designed to capture complex multimodal distributions, which misaligns with the limited complexity of the conditional traffic data distribution, causing models to underperform deterministic baselines.
To address this, we introduce the \underline{Co}nsistency-\underline{Re}laxed \underline{Flow} Map (CoReFlow) model, presenting the first effort to leverage the one-step generation capability of flow maps, known as solution operators of probability-flow ordinary differential equations. Besides, we improve accuracy using a concise training objective that supervises the model via interpolation endpoints over finite time intervals. Unlike existing methods (e.g., self-distillation and MeanFlow Identity) that maintain trajectory consistency to model multimodal distributions, our objective leverages conditional mean to restrict the consistency, thereby achieving higher accuracy in the probabilistic traffic forecasting task. Extensive experiments on four real-world datasets show CoReFlow achieves state-of-the-art forecasting accuracy. It surpasses recent baselines, provides competitive uncertainty modeling, and reduces inference costs compared to existing generative baselines.

\end{abstract}


\section{Introduction}

As a fundamental component of Intelligent Transportation Systems (ITS), traffic forecasting supports a wide range of downstream applications such as route planning, traffic signal control, and congestion management \cite{yin2021deep}. 
Most existing deep learning methods formulate traffic forecasting as a deterministic task and produce point predictions \cite{liu2023staeformer, dong2024heterogeneity, shao2025hyperd}.
However, transportation networks exhibit inherent uncertainty due to weather conditions, driver behaviors, and other external influences \cite{yan2023survey}, making probabilistic forecasting essential for reliable decision-making. 

\begin{figure}[t]
    \centering
    \includegraphics[width=\linewidth]{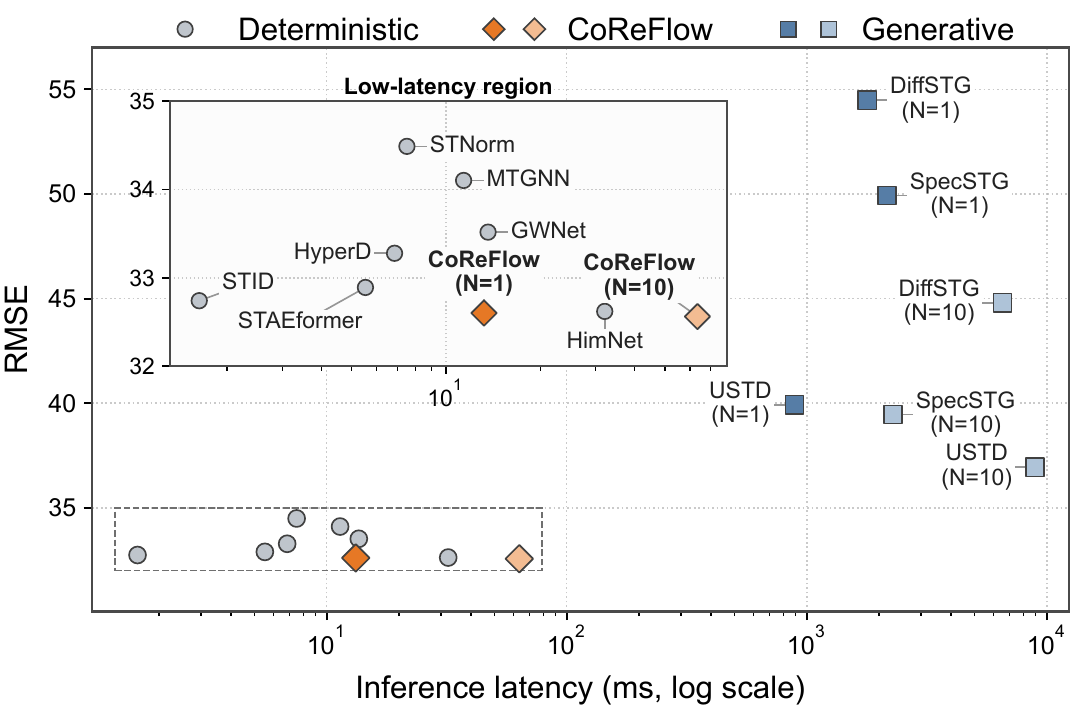}
    \caption{Accuracy (RMSE) vs. Inference latency on the PeMS07 dataset. The comparison includes CoReFlow, deterministic baselines, and generative baselines.}
    \label{fig:intro}
\end{figure}
Generative models, particularly Diffusion models \cite{ho2020denoising, song2021denoising}, have been increasingly applied to probabilistic traffic forecasting \cite{wen2023diffstg, chi2023difforecast, hu2024towards, lin2025specstg}. 
Conditioned on historical observations, these models generate future traffic states through learning to reverse a systematic noise-injection process.
Despite their generative capabilities, these models suffer from two major limitations: (1) {slow inference speed} due to an iterative, multi-step sampling process, and (2) a noticeable {performance gap} in point prediction accuracy compared to state-of-the-art deterministic baselines. These two limitations are illustrated in 
Figure~\ref{fig:intro}.


Our key observation is that the traffic data distribution tends to be concentrated, homogeneous, and of limited complexity when conditioned on priors such as specific time slots and weekdays. This suggests that the complex multimodal modeling capacity of diffusion models is largely redundant in this setting. Moreover, their multi-step sampling process may complicate optimization and introduce additional variance.


We address the probabilistic traffic forecasting problem by introducing the flow map, the solution operator of the probability-flow Ordinary Differential Equation (ODE), enabling efficient one-step generation. 
Specifically, this ODE is defined by a velocity field that transports a Gaussian prior to the target future traffic distribution, conditioned on historical traffic states and auxiliary variables.
Motivated by the aforementioned key observation, we propose a simple yet effective objective function that relaxes the trajectory consistency constraints for the flow map while simultaneously stabilizing the training process. 
Furthermore, we theoretically and empirically demonstrate the advantages of the objective. 
Based on the training objective, we propose the model CoReFlow, for \underline{Co}nsistency-\underline{Re}laxed \underline{Flow} Map, which demonstrates promising deterministic and stochastic performance, as illustrated in Figure~\ref{fig:intro}. The main contribution of this study can be summarized as:

\begin{itemize}
    \item CoReFlow is, to the best of our knowledge, the first model to apply the flow map concept to probabilistic traffic forecasting. It enables efficient one-step generation.
    \item Leveraging the limited complexity of conditional traffic distributions, we formulate a domain-specific training objective for CoReFlow. We provide a theoretical analysis of this objective and empirically show that it surpasses existing flow map training objectives.
    \item Extensive experiments on real-world public datasets show that CoReFlow achieves competitive performance in both deterministic and probabilistic forecasting.
\end{itemize}

\section{Method}

\subsection{Limited Complexity of Conditional Traffic Data Distributions}\label{sec:observation}
To analyze the impact of conditional information on different data distributions, we calculated the intra-class and inter-class cosine similarity distributions on ImageNet-256 and PEMS07, respectively. 
For the ImageNet-256 dataset, we randomly sampled 50 images from each of 150 randomly selected classes.
After normalization, each image is encoded using the pretrained VAE from Latent Diffusion Models~\cite{rombach2022high} and flattened.   
For PEMS07, the two time indices, "non-overlap-hour" and "day-of-the-week", were combined to generate $24 \times 7 = 168$ class labels. 
After applying Z-score normalization, the data was flattened into one-dimensional vectors. 
Given the differences in data dimension and processing methods between the two datasets, we avoid directly comparing the absolute similarity values across the two datasets.
Instead, we focus on the relative relationship between the intra-class and inter-class similarity distributions within each dataset.

\begin{figure}[t]
    \centering
    \includegraphics[width=\linewidth]{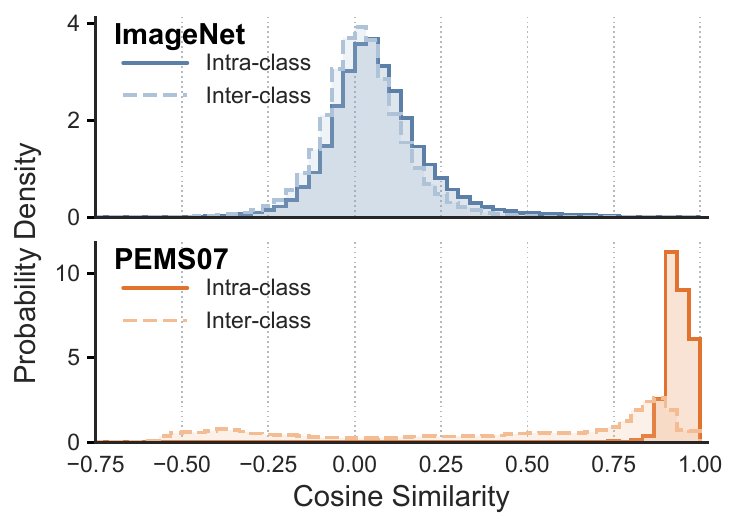}
    \caption{Distributions of intra- and inter-class cosine similarity on ImageNet and PEMS07.}
    \label{fig:method1}
\end{figure}

The similarity distribution is illustrated in Figure~\ref{fig:method1}. For ImageNet-256, the inter- and intra-class similarity distributions approximate a normal distribution centered at $0$, with significant overlap. This indicates that the conditional information imposes a weak constraint on the sample distribution.
The inter-class similarity distribution of PEMS07 exhibits a broad range from $-0.7$ to $1$.
This indicates non-uniform disparities across classes, which aligns with the temporal heterogeneity of traffic data highlighted in earlier studies \cite{jiang2023spatio, ji2023spatio, li2021spatial}.
Conversely, the intra-class similarity is densely concentrated in the range from $0.8$ to $1$, indicating a highly compact and homogeneous sample distribution. Furthermore, this suggests that the samples are tightly localized around the distribution mean.

The result demonstrates that auxiliary information such as time-of-day and day-of-week substantially reduces the complexity of the conditional distribution for traffic data. 
Consequently, the multi-step sampling process in diffusion models, which is typically advantageous for modeling complex, multimodal distributions, might not be fully utilized. 
Instead, the redundancy may introduce additional optimization difficulties and prediction variance, potentially explaining why diffusion models fail to consistently achieve an advantage in probabilistic traffic prediction.

\subsection{Traffic Forecasting via Flow Map}

For any given current time step, let $\bm x_h \in \mathbb{R}^{N \times T \times C}$ denote the historical traffic states collected over the preceding $T$ time steps, where $N$ is the number of spatial nodes and $C$ represents the feature dimension. 
Accordingly, the future traffic states to be predicted over the subsequent $T'$ time steps are denoted by $\bm x_f \in \mathbb{R}^{N \times T' \times C}$. 
Furthermore, we incorporate contextual priors via auxiliary variables: $\bm{c}_{\mathrm{tod}}, \bm{c}_{\mathrm{dow}} \in \mathbb{N}^{T}$ indicates the time-of-day and day-of-week across the $T$ historical steps, while $\bm{c}_{\mathrm{loc}} \in \mathbb{N}^{N}$ indicates spatial locations. 
For notational simplicity, we concatenate the historical traffic states and auxiliary variables into a single conditioning context: $\bm c = (\bm x_h, \bm c_\mathrm{tod}, \bm c_\mathrm{dow}, \bm c_\mathrm{loc})$.

We further define a marginal velocity field \cite{lipman2023flow, albergo2025stochastic}, denoted as $\bm v(\bm z_t, t\mid \bm c)$, that transports probability flow from a simple prior distribution $\mathcal{N}(\bm 0,\bm I)$ to the target future traffic distribution $p_\mathrm{data}(\bm x_f)$. This field is conditioned on historical traffic states and other auxiliary variables, collectively denoted as $\bm c = (\bm x_h, \bm c_\mathrm{tod}, \bm c_\mathrm{dow}, \bm c_\mathrm{loc})$ for notational simplicity.
The path for the probability flow is constructed
through a linear interpolation, 
$\bm z_t = t \bm x_f + (1-t)\bm \epsilon$, where $\bm{\epsilon} \sim \mathcal{N}(\bm{0}, \bm{I})\in \mathbb{R}^{ N\times T'\times C}$ is the prior sample. 

\begin{figure}[t]
    \centering
    \includegraphics[width=.95\linewidth]{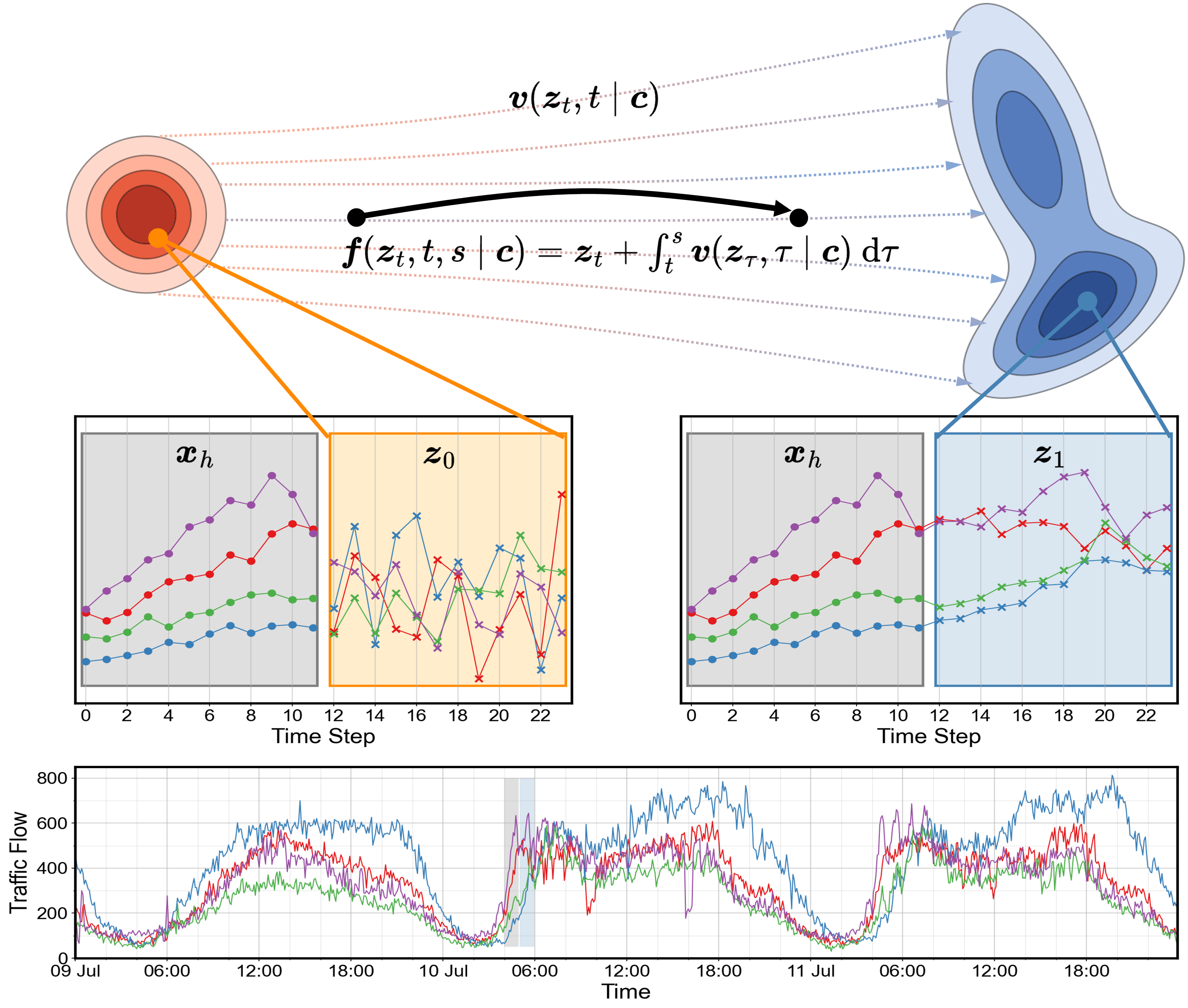}
    \caption{Velocity field and flow map defined in the traffic forecasting task.}
    \label{fig:method2}
\end{figure}

The flow map $\bm f$ is a solution operator of the ordinary differential equation (ODE) defined by the marginal velocity field, as shown in Figure~\ref{fig:method2}. 
It can be formulated via the integral equation on $\bm v(\bm z_t, t\mid \bm c)$:
\begin{equation}\label{eq:flow_map}
    \bm f(\bm z_t, t,s \mid \bm c) =\bm{z}_t + \int_t^s \bm{v}(\bm{z}_\tau, \tau\mid \bm c ) \, \mathrm{d}\tau,\quad s\ge t,
\end{equation} 
which naturally satisfies the boundary condition as well as the infinitesimal condition:
\begin{align}
    \bm f(\bm z_t,t,t\mid \bm c) &= \bm z_t \label{eq:boundary}\\
    \left. \tfrac{\partial}{\partial s} \bm f(\bm z_t, t, s \mid \bm c) \right|_{s=t}  &= \bm v(\bm z_t,t\mid \bm c) \label{eq:flow_matching}
\end{align}
    
In this study, we want to approximate the flow map using a neural network $\bm f_{\bm\theta}$ parameterized by $\bm{\theta}$, referred to as CoReFlow. To strictly satisfy the boundary condition in Equation~\ref{eq:boundary}, and in line with previous studies \cite{luo2025soflow, sabour2026align}, we parameterize $\bm{f}_{\bm{\theta}}$ as a residual mapping, formulated as the sum of the current state $\bm z_t$ and a residual network $\bm{F}_{\bm{\theta}}$:
\begin{equation}
\bm{f}_{\bm{\theta}}(\bm{z}_t, t, s\mid \bm c) = a(t, s) \bm{z}_t + b(t, s) \bm{F}_{\bm{\theta}} (\bm{z}_t, t, s\mid \bm c),
\end{equation}
where $a(t, s)$ and $b(t, s)$ are differentiable functions satisfying $a(t, t)=1$ and $b(t, t)=0$. 
We use the Euler parameterization \citep{sabour2026align, boffi2026build}, where,
\begin{equation}
    a(t,s) = 1,\quad b(t,s) = s-t.
\end{equation} 
This leads to a constant addition of Gaussian noise to the model output. 

During inference, this formulation enables highly efficient one-step generation, avoiding the need for iterative ODE solvers. By drawing an initial sample $\bm{z}_0=\bm \epsilon \sim\mathcal{N}(\bm 0, \bm I)$ and setting $t=0, s=1$, we directly map the prior to a generated data sample $\hat{\bm{x}}_f$ in a single forward pass as $\hat{\bm{x}}_f = \bm{z}_0 + \bm{F}_{\bm{\theta}}(\bm{z}_0, 0, 1 \mid \bm{c})$.

\subsection{Training Objective}

Training $\bm f_{\bm \theta}$ usually involves two losses~\citep{luo2025soflow, zhang2025alphaflow, frans2025one, boffi2026build}: Conditional Flow Matching loss $\ell_\mathrm{CFM}$ and Trajectory Consistency loss $\ell_\mathrm{TC}$. 
Specifically, $\ell_\mathrm{CFM}$ optimizes the model by matching its diagonal derivative, as defined in Equation~\ref{eq:flow_matching}, to conditional velocity targets constructed from stochastic interpolations. 
While $\ell_\mathrm{CFM}$ only constrains $\bm f_{\bm \theta}$ on the diagonal ($s=t$), $\ell_\mathrm{TC}$ regularizes off-diagonal ($s>t$) behaviors via self-consistency along the ODE trajectory, thereby enabling one-step generation of multimodal distributions without collapsing to the mean~\cite{song2023consistency, frans2025one}.
Nevertheless, incorporating $\ell_\mathrm{TC}$ inevitably introduces optimization difficulties, such as gradient instability \cite{boffi2025flow}, gradient incompatibility with $\ell_\mathrm{CTM}$ \cite{frans2025one, zhang2025alphaflow}, and error accumulation during early training stage  \cite{song2024improved, boffi2025flow}.

Faced with the issues arising from $\ell_\mathrm{TC}$, we exploit the inherently low-complexity and compactness of the conditional traffic data distribution, as noted in Section~\ref{sec:observation}. By relaxing the trajectory consistency in the training objective, we simplify the optimization process while preserving training stability. Specifically, 
Specifically, we propose a consistency-relaxed loss ($\ell_{\mathrm{CR}}$) that leverages the interpolation path to guide the training of the model $f_{\bm{\theta}}$, defined as:
\begin{equation}\label{eq:cr_loss}
    \ell_{\mathrm{CR}}(\bm{\theta})=\mathbb{E}_{\bm{x}_f, \bm{\epsilon}, 
    \bm c, t, s} \left[ \tfrac{1}{(s-t)^2}\left\|  \bm f_{\bm{\theta}}(\bm z_t , t, s\mid \bm c) - \bm z_s \right\|^2_2 \right],
\end{equation}
where  $\bm z_t$ and $\bm z_s$ are constructed via the interpolation $\bm z_k = k \bm{x}_f + (1-k) \bm{\epsilon}$ for $k \in \{t, s\}$, with $s > t$. 
To theoretically justify this relaxation, we first analyze its approximation behavior to ensure that dropping strict consistency does not degrade the target flow map.
\begin{proposition}[Approximation Error, Proof in supplementary material]\label{theorem:error_bound}
    The optimal minimizer for $\ell_{\mathrm{CR}}(\bm\theta)$ approximates the flow map $\bm f$ with an error of $\mathcal{O}(|s-t|^2)$.
\end{proposition}
Beyond approximation guarantees, our objective formulation establishes a connection between trajectory consistency training and Conditional Flow Matching.
\begin{proposition}[Convergence to CFM Loss, Proof in supplementary material] \label{theorem:cfm_reduction}
    Under the limit as $s \rightarrow t$, $\ell_{\mathrm{CR}}$ reduces to the Conditional Flow Matching objective, i.e., $\lim_{s \rightarrow t} \ell_{\mathrm{CR}}(\bm{\theta}) = \ell_{\mathrm{CFM}}(\bm{\theta})$.
\end{proposition}

\subsubsection{Node-Level Adapted Loss Weight }
Previous studies  \cite{geng2025consistency, luo2025soflow,geng2026mean} demonstrate that 
incorporating an adapted weighting mechanism into the loss function helps to stabilize training dynamics.
The objective function for the flow map $\bm f_{\bm \theta}$ in Equation~\ref{eq:cr_loss}  can be reformulated as $\ell_\mathrm{CR}(\bm\theta) = \mathbb{E}_{\bm x_f, \bm \epsilon, \bm c, t, s} [ \left\| \bm \Delta \right\|^2_2 ]$, where,
\begin{equation}
\bm\Delta = \frac{1}{(s-t)}\left(\bm f_{\bm{\theta}}(\bm z_t , t, s\mid \bm c) - \bm z_s\right)\in \mathbb{R}^{ N\times T^\prime \times C}. 
\end{equation}
Rather than assigning an adaptive weight to the loss on the sample level \cite{luo2025soflow,geng2026mean}, we propose an adaptive weight at the node level.
Specifically, given the residual $\bm\Delta$,  the adaptive weight $w_n$ for $n$-th node is based on the sum of squared errors on its temporal axis,
\begin{equation}
\begin{aligned}
    e_n = \sum\nolimits_{j,k} \bm\Delta_{n,j,k}^2 \ ,\quad
    w_n = \frac{1}{\left( \mathrm{sg}\left( e_n\right) + c \right)^p}\ ,
\end{aligned}
\end{equation}
where $\mathrm{sg}(\cdot)$ denotes the stop-gradient operation, $c$ is a small constant for numerical stability, and $p$ is a scaling hyperparameter. The final optimized objective is given by computing the mean of the adaptively weighted errors across all nodes:
\begin{equation}
    \ell^\mathrm{adp}_{\mathrm{CR}}(\bm\theta) = \frac{1}{N} \sum_{n=1}^{N} w_ne_n \ .
\end{equation}

\subsection{Model Structure and Conditioning}
\begin{figure}[t]
    \centering
    \includegraphics[width=\linewidth]{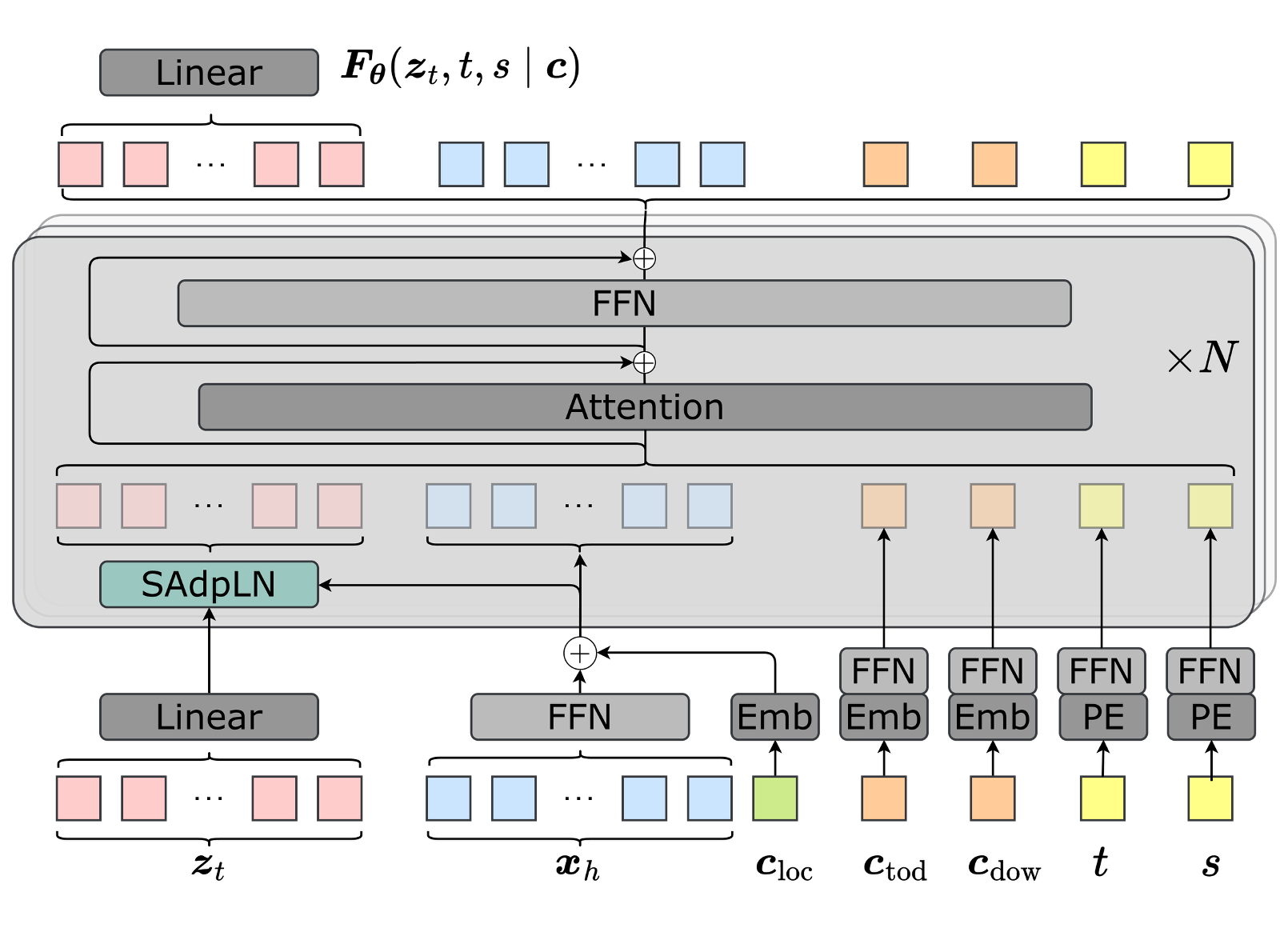}
    \caption{The Overall Model Structure.}
    \label{fig:model_structure}
\end{figure}

We employ a Transformer \cite{vaswani2017attention} as the backbone of our model. Given the dimension-wise correspondence between the conditional information $\bm c$ and the input signal $\bm z_t$, 
we find structure-level conditioning much more intuitive and suitable than guidance-based conditioning like classifier-free guidance (CFG) \cite{ho2021classifierfree}.
Thus, our model structure, as shown in Figure~\ref{fig:model_structure}, is designed to directly inject and fuse the conditions into the input.

\subsubsection{Data Embeddings} 
To effectively handle both continuous traffic data and discrete contextual attributes, we propose a specialized feature encoding module.
For clarity, we first define a standard feedforward network ($\mathrm{FFN}$) consisting of two linear transformations separated by a GELU activation function \cite{hendrycks2017bridging}. For a generic input $\bm x$, the $\mathrm{FFN}$ is defined as:
$$ \mathrm{FFN}(\bm x) = \mathrm{GELU}(\bm x \bm w_1 + \bm b_1) \bm w_2 + \bm b_2, $$
where $\bm w_1, \bm w_2$ are learnable weight matrices and $\bm b_1, \bm b_2$ are bias terms. 
The historical traffic states $\bm x_h$ is first flatten from $\mathbb{R}^{N\times T\times C}$ to $\mathbb{R}^{N\times (T \cdot C)}$ to merge the temporal and channel dimensions, and subsequently projected to a $ D$-dimensional representation $\bm h_x$ via an FFN,
\begin{equation}
    \bm h_x = \mathrm{FFN}(\mathrm{Reshape}({\bm x}_h)) \in \mathbb{R}^{N\times D}.
\end{equation}
Similarly, the flow path variable $\bm z_t$ is flatten from  $\mathbb{R}^{ N\times T^\prime\times C}$ to $\mathbb{R}^{N\times (T^\prime \cdot C)}$ and mapped to $D$-dimensional latent space via a linear projection,
\begin{equation}
    \bm h_z = \mathrm{Linear}(\mathrm{Reshape}({\bm z}_t)) \in \mathbb{R}^{N\times D}.
\end{equation}
Besides, categorical variables, $\bm c_\mathrm{tod}$, $\bm c_\mathrm{dow}$, and $\bm c_\mathrm{loc}$, are mapped into $D$-dimensional latent spaces using learnable embedding matrices. Mathematically, this is equivalent to applying a linear transformation to their one-hot representations:
\begin{equation}
    \bm e_k = \text{OneHot}(\bm c_k) \bm w_k, \quad k \in \{\mathrm{tod}, \mathrm{dow}, \mathrm{loc}\}
\end{equation}
where $\bm w_{\mathrm{tod}} \in \mathbb{R}^{V_{\mathrm{tod}} \times D}$, $\bm w_{\mathrm{dow}} \in \mathbb{R}^{V_{\mathrm{dow}} \times D}$, and $\bm w_{\mathrm{loc}} \in \mathbb{R}^{N \times D}$ represent the trainable parameters, with $V_\mathrm{tod}$ and $V_\mathrm{dow}$ being the number of unique categories. The resulting embeddings are dimensioned as $\bm e_{\mathrm{tod}}, \bm e_{\mathrm{dow}} \in \mathbb{R}^{T\times D}$ and $\quad \bm e_{\mathrm{loc}} \in \mathbb{R}^{N\times D}$.
Furthermore, the location embedding $\bm e_\mathrm{loc}$ is  and added to $\bm h_x$, element-wisely:
\begin{equation}
    {\bm h}_{x} = \bm h_x + \bm e_\mathrm{loc} \ \in \mathbb{R}^{N\times D}.
\end{equation}
Temporal embeddings, $\bm e_\mathrm{tod}$ and $\bm e_\mathrm{dow}$, are averaged along the temporal dimension and nonlinearly projected via two FFNs respectively,
\begin{equation}
\begin{aligned}
    \bm h_\mathrm{tod} &= \mathrm{FFN}\left(\mathrm{Mean}(\bm e_\mathrm{tod})\right) \in \mathbb{R}^{1\times D},\\
    \bm h_\mathrm{dow} &= \mathrm{FFN}\left(\mathrm{Mean}(\bm e_\mathrm{dow})\right) \in \mathbb{R}^{1\times D},
\end{aligned}
\end{equation}
where $\mathrm{Mean}(\cdot)$ denotes the average pooling operation over the temporal axis $T$.

The time variables $t$ and $s$ are encoded by sinusoidal positional encoding, denoted as $\mathrm{PE}(\cdot)$, and projected to $\bm h_{t}$ and $h_{s}$ via FFNs, respectively,
\begin{equation}
\begin{aligned}
    \bm h_{k} =\mathrm{FFN}( \mathrm{PE}(k) ) \in \mathbb{R}^{1\times D},\quad k\in\{s,t\}\\
\end{aligned}
\end{equation}

Overall, the data embedding module projects the input variables
$\bm z_t$, $t$, $s$, and $\bm c$ into a $D$-dimensional latent space, with $ \bm h_z,{\bm h}_{x}\in \mathbb{R}^{N\times D}$ and $\bm h_{\mathrm{tod}},\bm h_{\mathrm{dow}}, \bm h_{t}, \bm h_{s} \in \mathbb{R}^{1\times D}$.

\subsubsection{Transformer with Simplified AdaLN}
The model utilizes a stack of Transformer blocks, each with a Simplified Adaptive Layer Normalization (SAdaLN) module. Let $l \in \{1,\ldots,L\}$ denote the block index, where $L$ is the total number of Transformer blocks. 
For the initial layer ($l=1$), the individual input representations are initialized as $\bm{h}^{(1)}_k = \bm{h}_k,\  \forall k \in \{z, x, \mathrm{tod}, \mathrm{dow}, t, s\}$, and the complete input sequence is formed by their concatenation $[\bm{h}^{(1)}_z \parallel \bm{h}^{(1)}_{x} \parallel \ldots \parallel \bm{h}^{(1)}_{s}]$.

Within the $l$-th block, the conditioning representation of historical traffic and location, $\bm h_{x}^{(l)}$, is modulated into the flow-path representation, $\bm h_{z}^{(l)}$, via SAdaLN. Specifically, the conditioning representation $\bm h_{x}^{(l)}$ is projected and split along the feature dimension to generate a shift parameter $\bm{\mu}^{(l)}$ and a scale parameter $\bm{\sigma}^{(l)}$:
\begin{equation}
\begin{aligned}
    [ \bm{\mu}^{(l)}, \bm{\sigma}^{(l)} ] &= \mathrm{FFN}(\bm h_{x}^{(l)}) \in \mathbb{R}^{N\times 2D},\\
    \tilde{\bm h}_z^{(l)} &= \bm h_z^{(l)} \odot \bm{\sigma}^{(l)} + \bm{\mu}^{(l)} \in \mathbb{R}^{N\times D},
\end{aligned}
\end{equation}
where $\odot$ denotes element-wise multiplication. 
The resulting joint sequence is then processed by the $l$-th Transformer block to capture dependencies among the flow-path, historical, temporal, and path indicator tokens:
\begin{equation}
    \begin{aligned}
    &[ \bm h_z^{(l+1)} \parallel \bm h_{x}^{(l+1)} \parallel \bm h^{(l+1)}_{\mathrm{tod},\mathrm{dow}, t,s} ] \\
    & = \mathrm{Transformer} ( [ \tilde{\bm h}_z^{(l)} \parallel \bm h_{x}^{(l)} \parallel \bm h^{(l)}_{\mathrm{tod},\mathrm{dow}, t,s} ] ),
\end{aligned}
\end{equation}
where $\bm h^{(l)}_{\mathrm{tod},\mathrm{dow}, t,s}= [\bm h_{\mathrm{tod}}^{(l)} \parallel \bm h_{\mathrm{dow}}^{(l)} \parallel \bm h_{t}^{(l)}\parallel \bm h_{t}^{(l)}]$.

\subsubsection{Output Layer}
The final flow-path representation $\bm h_z^{(L)} \in \mathbb{R}^{N\times D}$ from the Transformer output is linearly projected to $\mathbb{R}^{N \times (T^\prime\cdot C)}$ and reshaped to match the target dimensions, 
\begin{equation}
    \hat{\bm z}_s=\mathrm{Reshape}(\mathrm{Linear}(\bm h_z^{(L)})) \in \mathbb{R}^{ N\times T^\prime\times C}.
\end{equation}

\section{Experiments}
\subsection{Experimental Setup}
\subsubsection{Datasets}
A thorough evaluation of our model is conducted under extensive experiments on four real-world datasets: PEMS03, PEMS04, PEMS07, and PEMS08. The time resolution for the four datasets is 5 minutes, resulting in 12 time steps in one hour. More details are shown in Table~\ref{tab:dataset}.
\begin{table}[htbp]
    \centering
    \begin{tabular}{cccc}
    \toprule
    Datasets& \#Nodes & \#Timesteps & Time Range \\
    \midrule
    PEMS03   & 358  & 26,209 & 09/2018-11/2018\\
    PEMS04   &  307 & 16,992  & 01/2018-02/2018\\
    PEMS07 &883 & 28,224 & 05/2017-08/2017\\
    PEMS08  & 170 & 17,856 & 07/2016-08/2016\\
    \bottomrule
    \end{tabular}
    \caption{Dataset Description}\label{tab:dataset}
\end{table}

\subsubsection{Baselines}
We compare our model with 11 baseline models, which belong to two categories:
\begin{itemize}
    \item Deterministic models: HI \cite{cui2021historical}, GWNet \cite{wu2019graph}, MTGNN \cite{wu2020connecting}, STNorm \cite{deng2021st}, STID \cite{shao2022spatial}, STAEFormer \cite{liu2023staeformer}, HimNet \cite{dong2024heterogeneity}, and HyperD \cite{shao2025hyperd}.
    \item Generative models: DiffSTG \cite{wen2023diffstg}, USTD \cite{hu2024towards}, SpecSTG \cite{lin2025specstg}.
\end{itemize} 

\subsubsection{Implementation Details}
Four datasets are chronologically split into training, validation, and test sets in a 6:2:2 ratio, following the baseline protocols. 
Observations from the past hour ($T=12$) are used to predict traffic states for the next hour ($T'=12$).
Hyperparameters are selected through Bayesian optimization with 30 trials per dataset. The selected values and search space are reported in the supplementary material. All experiments are conducted on an NVIDIA GH200 GPU with 120\,GB of memory. 
Our code is provided in the supplementary material.

\begin{table*}[!t]
\centering
\renewcommand{\arraystretch}{1.2}
\small
\begin{tabular}{c|ccc|ccc|ccc|ccc}
\toprule
Model
& \multicolumn{3}{c|}{PEMS03}
& \multicolumn{3}{c|}{PEMS04}
& \multicolumn{3}{c|}{PEMS07}
& \multicolumn{3}{c}{PEMS08} \\
\cmidrule(lr){2-4}
\cmidrule(lr){5-7}
\cmidrule(lr){8-10}
\cmidrule(lr){11-13}
& RMSE & MAE & MAPE
& RMSE & MAE & MAPE
& RMSE & MAE & MAPE
& RMSE & MAE & MAPE \\
\midrule

HI
& 49.892 & 32.619 & 30.596
& 61.659 & 42.353 & 29.922
& 71.340 & 49.288 & 22.753
& 50.452 & 34.661 & 21.631 \\

GWNet
& \underline{25.080} & \underline{14.545} & 14.857
& 30.310 & 18.948 & 13.469
& 33.517 & 20.472 & 8.683
& 23.398 & 14.586 & 9.659 \\

MTGNN
& 25.653 & 14.953 & 15.523
& 30.940 & 19.084 & 13.407
& 34.102 & 20.958 & 8.985
& 24.281 & 15.252 & 10.001 \\

STNorm
& 26.129 & 15.275 & \underline{14.817}
& 31.440 & 19.056 & 12.861
& 34.488 & 20.508 & 8.639
& 24.860 & 15.406 & 9.838 \\

STID
& 27.606 & 15.298 & 16.166
& \underline{29.909} & 18.402 & 12.573
& 32.741 & 19.645 & 8.314
& 23.350 & 14.226 & 9.356 \\

STAEFormer
& 25.549 & 15.010 & 15.415
& 30.199 & \underline{18.154} & 12.455
& 32.891 & 19.438 & \underline{8.028}
& \underline{23.225} & \underline{13.511} & \underline{8.858} \\

HimNet
& 26.898 & 15.182 & 15.730
& 30.077 & 18.234 & \underline{12.076}
& \underline{32.620} & \underline{19.238} & 8.276
& 23.322 & 13.592 & 8.987 \\

HyperD
& \textbf{24.205} & 14.940 & 16.023
& 30.018 & 18.240 & 12.510
& 33.278 & 19.434 & 8.116
& 23.318 & 13.640 & 9.016 \\

\midrule

\textbf{CoReFlow}
& 25.321 & \textbf{14.460} & \textbf{14.389}
& \textbf{29.880} & \textbf{17.936} & \textbf{11.774}
& \textbf{32.601} & \textbf{18.783} & \textbf{7.752}
& \textbf{22.875} & \textbf{13.197} & \textbf{8.581} \\

\bottomrule
\end{tabular}
\caption{
Performance comparison with deterministic models.
CoReFlow is evaluated based on 1-sample generation ($N{=}1$).
All results are averaged over three runs with random seeds.
Best and second-best results are shown in bold and underlined, respectively.
}
\label{tab:performance_comparison_det}
\end{table*}

\subsubsection{Evaluation Metrics} 
For deterministic models,  three metrics are used: (1) Mean Absolute Error (MAE), (2) Mean Absolute Percentage Error (MAPE), and (3) Root Mean Squared Error (RMSE). For generative models, point metrics are computed from the predictive sample mean and the Continuous Ranked Probability Score (CRPS) is used as an additional metric for evaluating probabilistic prediction. In this study, CRPS is calculated using the Monte Carlo method \citep{gneiting2007strictly}.

\subsection{Performance Comparison}

\subsubsection{Deterministic Predictions}
Table \ref{tab:performance_comparison_det} compares CoReFlow with eight deterministic baselines across four datasets. Notably, CoReFlow ranks first in 11 of the 12 evaluated cases, outperforming baselines on all metrics for PEMS04, PEMS07, and PEMS08, as well as on MAE and MAPE for PEMS03.
The only exception is RMSE on PEMS03, where HyperD ranks first.
On average across the four datasets, CoReFlow reduces the MAPE by approximately $3.0\%$ compared to the strongest baseline, reaching a maximum reduction of $3.4\%$ on PEMS07.

Overall, these results demonstrate that CoReFlow maintains competitive performance on deterministic metrics, even when generating only a single sample.

\subsubsection{Generative Predictions}

\begin{table}[t]
    \centering
    \small
    {
    \renewcommand{\arraystretch}{1.2}
    \setlength{\tabcolsep}{5pt}
    \begin{tabular}{c|cc|ccc|c}
        \toprule
        \multicolumn{3}{c}{Metric}
        & DiffSTG & USTD & SpecSTG & \textbf{CoReFlow} \\
        \midrule

        \multirow{7}{*}{\rotatebox{90}{PEMS07}}
        & \multirow{3}{*}{\rotatebox{90}{1 sample}}
        & RMSE & 55.926 & \underline{40.747} & 49.917 & \textbf{32.601} \\
        & & MAE  & 39.047 & \underline{26.602} & 35.630 & \textbf{18.783} \\
        & & MAPE & 21.431 & \underline{12.699} & 19.172 & \textbf{7.752} \\
        \cmidrule(lr){2-7}
        & \multirow{4}{*}{\rotatebox{90}{10 samples}}
        & RMSE & 44.802 & \underline{36.934} & 39.458 & \textbf{32.562} \\
        & & MAE  & 31.085 & \underline{24.297} & 26.298 & \textbf{18.731} \\
        & & MAPE & 18.430 & \underline{12.060} & 13.759 & \textbf{7.693} \\
        & & CRPS & 21.335 & 19.226 & \underline{18.527} & \textbf{18.201} \\
        \midrule

        \multirow{7}{*}{\rotatebox{90}{PEMS08}}
        & \multirow{3}{*}{\rotatebox{90}{1 sample}}
        & RMSE & 34.577 & \underline{30.707} & 34.045 & \textbf{22.875} \\
        & & MAE  & 22.921 & \underline{19.801} & 24.370 & \textbf{13.197} \\
        & & MAPE & 14.561 & \underline{13.801} & 21.507 & \textbf{8.581} \\
        \cmidrule(lr){2-7}
        & \multirow{4}{*}{\rotatebox{90}{10 samples}}
        & RMSE & 27.334 & \underline{26.892} & 27.204 & \textbf{22.843} \\
        & & MAE  & 17.921 & \underline{17.456} & 18.175 & \textbf{13.163} \\
        & & MAPE & \underline{11.511} & 12.780 & 16.106 & \textbf{8.546} \\
        & & CRPS & 14.016 & 13.161 & \underline{12.817} & \textbf{12.741} \\
        \bottomrule
    \end{tabular}
    }
    \caption{
    Performance comparison with generative models. Best and second-best results are shown in bold and underlined, respectively.}\label{tab:performance_comparison_gen_mean}
\end{table}

Table~\ref{tab:performance_comparison_gen_mean} compares CoReFlow with three generative baselines on PEMS07 and PEMS08, which are the largest and smallest road networks in our benchmark with 883 and 170 sensors, respectively.
It achieves the best result in all 14 reported comparisons.
Under the 1-sample generation setting, it reduces RMSE, MAE, and MAPE by $120.0\%$, $29.4\%$, and $39.0\%$, respectively, compared to the strongest baseline on PEMS07.

Under the 10-sample aggregation setting, our model maintains the lowest RMSE, MAE, and MAPE, as well as achieving the best CRPS across both datasets.
The joint improvements in deterministic and probabilistic metrics demonstrate that CoReFlow generates predictive distributions of superior accuracy compared to existing generative baselines.

\subsection{Ablation Study}
This ablation study mainly aims to validate the design of the training objective $\ell_\mathrm{CR}$ for our model.
We compare our approach with recently proposed flow map training methods, including MeanFlow Identity \cite{geng2026mean}, improved MeanFlow Identity \cite{geng2025improved}, and self-Distillation \cite{boffi2026build}.
Additionally, we substitute our adapted loss weight from the node level to the batch level for comparison. Since these losses were originally designed for image datasets, we reproduced them as faithfully as possible. 
We restrict this analysis to PEMS07 and PEMS08, as they represent the largest and smallest datasets, respectively. 
For notation clarity, the necessary formulation is established first.
\begin{align}
    \dot{\bm z}_t &= \tfrac{\mathrm{d}}{\mathrm{d}t}\bm z_t =  \bm x_f - \bm\epsilon\\
    \bm v_{\bm\theta}^{t,\bm c}(\bm z) &:= \left.\tfrac{\partial}{\partial s} \bm f_{\bm{\theta}}(\bm z, t, s\mid \bm c) \right|_{s=t}\\
    \bm f_{\bm\theta}^{t,s,\bm c}(\bm z) &:= \bm f_{\bm\theta}(\bm z,t,s\mid\bm c)\\
    \bm F_{\bm\theta}^{t,s,\bm c}(\bm z) &:= \bm F_{\bm\theta}(\bm z,t,s\mid\bm c)
\end{align}

\subsubsection{Self-Distillation} Self-distillation \cite{boffi2026build} is a training objective, $\ell_\mathrm{sd}$, combining the Conditional Flow Matching loss as well as the Trajectory Consistency loss inspired by the Lagrangian condition. The formulation is:
\begin{equation}
\begin{aligned}
    \ell_\mathrm{sd}(\bm\theta) &= \ell_\mathrm{CFM}(\bm\theta) + \ell_\mathrm{TC}(\bm\theta)\\
    \ell_\mathrm{CFM}(\bm\theta)  &= \mathbb{E}_{ \bm x_f, \bm \epsilon, t}  [\|  \bm v_{\bm\theta}^{t,\bm c}(\bm z_t)  - \dot{\bm z}_t  \|_2^2 ],\\
    \ell_\mathrm{TC}(\bm\theta) &= \mathbb{E}_{\bm x_f, \bm \epsilon, t,s} [\|\tfrac{\partial}{\partial s}\bm f_{\bm\theta}^{t,s,\bm c}(\bm z_t)  - \bm g_{\bm\theta^-}^{t,s,\bm c}(\bm z_t) \|_2^2 ],\\
\end{aligned}
\end{equation}
where $\bm g_{\bm\theta^-}^{t,s,\bm c}(\bm z_t) =\bm v_{\bm\theta^-}^{t,\bm c}(\bm f_{\bm\theta^-}^{t,s,\bm c}(\bm z_t))$ and $\bm\theta^-$ means a stop-gradient operation on $\bm\theta$. Besides, the norm in $\ell_\mathrm{CFM}$ and $\ell_\mathrm{TC}$ is weighted by a trainable model $w_{\bm\theta}(t,s)$ in the form of $e^{w_{\bm\theta}(t,s)}\|\bm\Delta \|^2_2 - w_{\bm\theta}(t,s)$.

\subsubsection{MeanFlow Identity}
MeanFlow~\cite{geng2026mean} Identity is an objective $\ell_\mathrm{mf}$ formulated as:
\begin{equation}\label{eq:mf_id}
\begin{aligned}
    \ell_\mathrm{mf}(\bm\theta) &= \mathbb{E}_{\bm x_f, \bm \epsilon, t,s} [\| \bm F_{\bm\theta}^{t,s,\bm c}(\bm z_t) -  \dot{\bm z}_t  +\bm F_{\mathrm{tgt}}\|^2_2],\\
    \bm F_{\mathrm{tgt}} &=  (s-t) ( \dot{\bm z}_t \tfrac{\partial}{\partial \bm z_t} \bm F_{\bm\theta^-}^{t,s,\bm c}(\bm z_t)+\tfrac{\partial}{\partial t} \bm F_{\bm\theta^-}^{t,s,\bm c}(\bm z_t)).
\end{aligned}
\end{equation}
Batch-level adapted loss weight, $w=1/(\|\bm\Delta\|^2_2+c)^p$ is applied to the norm of $\ell_\mathrm{mf}$.

\subsubsection{Improved MeanFlow Identity}
Improved MeanFlow Identity~\cite{geng2025improved} is the the a training objective $\ell_\mathrm{imf}$ based on  $\ell_\mathrm{mf}$. Specifically, the $\dot{\bm z}_t$ in $\bm F_{\mathrm{tgt}}$ in Equation~\ref{eq:mf_id} is replaced by $\bm v_{\bm \theta}^{t,\bm c}(\bm z_t)$.

\subsubsection{Adaptive Loss Weight}
We compared the node-level adaptive loss weight to the batch-level alternative
based on the square L2 loss, $w=1/(\|\bm\Delta\|^2_2+c)^p$
\cite{geng2025consistency, geng2026mean}.

\subsubsection{Summary}
\begin{figure}[t]
    \centering
    \includegraphics[width=\linewidth]{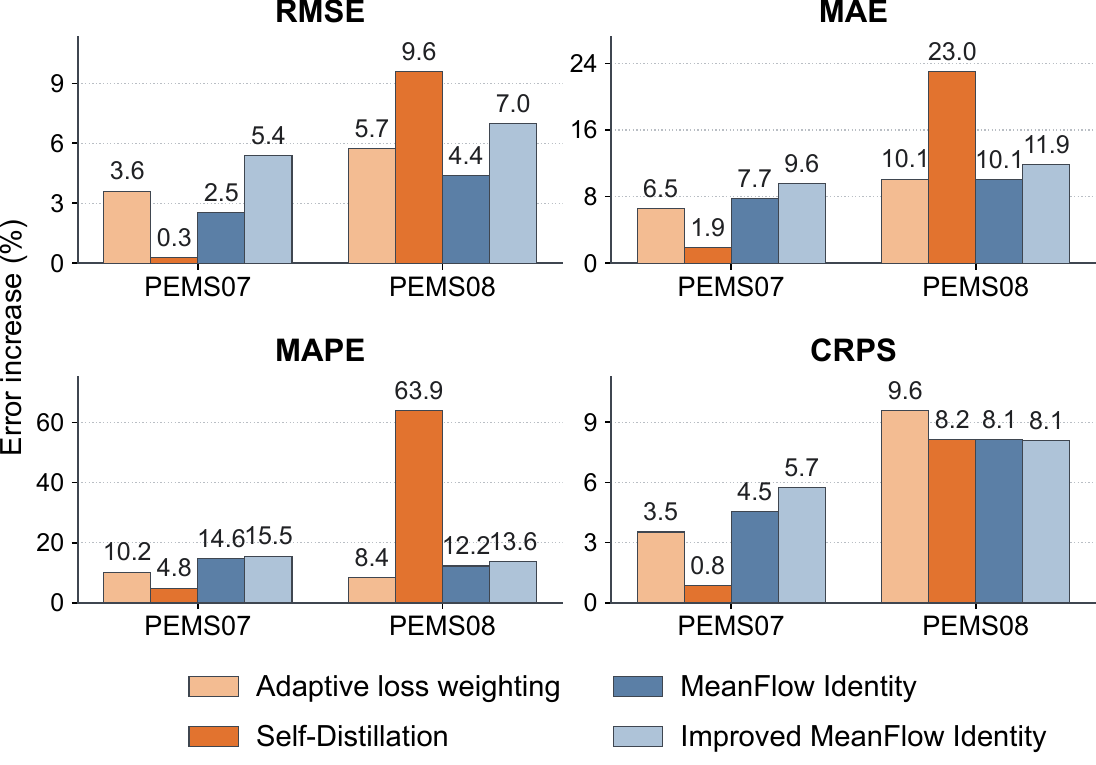}
    \caption{Percentage increase in prediction error for different ablated
    variants on the PEMS07 and PEMS08 datasets, using the proposed CoReFlow
    model as the baseline.
    }
    \label{fig:ablation}
\end{figure}
Figure~\ref{fig:ablation} summarizes the results of the ablation experiments by illustrating the relative percentage increase in prediction error for each
variant. The positive error increases across all metrics clearly demonstrate
that the current flow map training objectives degrade the model's predictive capability. Replacing the  $\ell_{\mathrm{CR}}$ with the other 
objective compromises model accuracy, particularly on the PEMS08 dataset,
highlighting the effectiveness of the proposed training objective. Second, training with a batch-level adaptive loss weight consistently underperforms, verifying that node-level adaptive weighting is essential for effective training.

\subsection{Inference Efficiency}

\begin{figure}[t]
    \centering
    \includegraphics[width=\columnwidth]{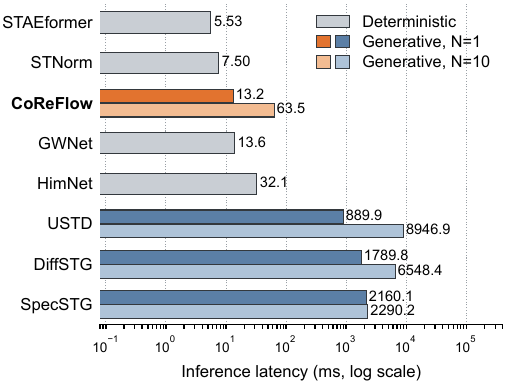}
    \caption{
    Inference latency on PEMS07.}
    \label{fig:inference_latency}
\end{figure}

We measure the inference latency for generating the complete 12-step forecast with $S\in\{1,10\}$ trajectories. The efficiency gain follows directly from the sampling formulation. CoReFlow evaluates the endpoint flow map once to generate the complete forecast. It therefore requires one network function evaluation (NFE) per trajectory~\cite{chen2018neural}. In contrast, SpecSTG requires 600 NFEs because it performs 50 diffusion evaluations in each of 12 autoregressive forecasting rounds. The reduced sampling depth removes the serial network evaluations required by iterative generative models.

As shown in Fig.~\ref{fig:inference_latency}, CoReFlow achieves over an order-of-magnitude faster inference than generative baselines on PEMS07, the largest dataset. It achieves speedups ranging from $67\times$ to $163\times$ for $S=1$, and from $36\times$ to $141\times$ for $S=10$, over the generative baselines. Its single-trajectory latency is comparable to Graph-based models (GWNet and HimNet), although deterministic baselines remain faster. The same efficiency advantage is observed on PEMS08. The full protocol and results is given in the supplementary material.


\section{Related Work}

\subsection{Deterministic Spatiotemporal Forecasting}
Early spatiotemporal models couple graph neural networks with temporal convolutions to capture spatial and temporal dependencies, either by learning an adaptive adjacency matrix \citep{wu2019graph} or by learning the graph structure jointly with the forecasting objective \citep{wu2020connecting}. Recent work shows that graph modules are not essential, reaching competitive accuracy through spatial and temporal normalization \citep{deng2021st}, identity embeddings on a multilayer perceptron \citep{shao2022spatial}, or an adaptive embedding in a Transformer backbone \citep{liu2023staeformer}. Later models further exploit the heterogeneity and periodicity of traffic \citep{dong2024heterogeneity,shao2025hyperd}. These models predict accurately, but they output a single trajectory and cannot express the uncertainty required for risk-aware decisions.

\subsection{Generative Probabilistic Forecasting}
Generative models instead sample future states from a learned conditional distribution, and most such models for spatiotemporal graphs are built on diffusion. The idea began in probabilistic time series forecasting \cite{rasul2021autoregressive,tashiro2021csdi} and was carried to spatiotemporal graphs by DiffSTG \citep{wen2023diffstg}. Later models mainly differ in how past observations and spatial structure enter the denoiser, through imputation-style masking \citep{liu2023pristi}, a pretrained deterministic encoder \citep{hu2024towards}, or the graph spectral domain \citep{lin2025specstg}. Flow matching offers an alternative that transports the prior to the data distribution without a diffusion noise schedule \citep{kollovieh2025flow}. Both paradigms capture uncertainty at two recurring costs. Sampling runs the network many times in sequence and is slow, and the point accuracy still falls behind the strongest deterministic models.

\subsection{One-Step Generation and Flow Maps}
Both diffusion and flow matching define generation as an ordinary differential equation solved with many small steps. One-step generators remove this cost by learning its solution operator directly, which is the flow map \citep{boffi2025flow}. Consistency models are an early instance that maps every point on a trajectory to the same endpoint \citep{song2023consistency}, and later work generalizes this to arbitrary step sizes and continuous time \citep{kim2024consistency,frans2025one,sabour2026align,zhang2025alphaflow}, including MeanFlow, which learns the average velocity over an interval so one step spans the whole path \citep{geng2026mean}. A unifying analysis views these methods as approximations of the same solution operator \citep{boffi2026build}, and most of them add a self-consistency term that keeps distinct modes apart, which matters for strongly multimodal image data.


\section{Conclusion}
We studied probabilistic traffic forecasting under a flow map formulation. We observed that traffic futures concentrate around a dominant mode given recent observations. This observation motivates relaxing trajectory consistency. Guided by this insight, we proposed a consistency-relaxed surrogate objective with a bounded approximation error. We then instantiated the objective in a Transformer flow map that produces a forecast in a single network evaluation. Experiments on four datasets showed accurate point predictions and competitive probabilistic scores. Both were achieved at a small fraction of the sampling cost of iterative baselines.

Our study has two main limitations. First, the proposed objective trades a small approximation bias for stable supervision, so its benefit is largest when the conditional future is close to unimodal, which is the regime our datasets fall in. Second, our evaluation covers highway flow datasets with a one-hour horizon, and other domains may exhibit stronger multimodality. Future work will extend the relaxed flow map to settings with richer multimodality and longer horizons.

\clearpage
\bibliography{aaai2027}


\section{Proofs}
\label{app:proofs}

\subsection{Proof of Proposition 1}

\begin{proof}
The objective function $\ell_{\mathrm{CR}}$ is defined as an expected squared $L_2$ distance. For any fixed time steps $t$ and $s$, and conditioning variable $\bm c$, minimizing the expected loss over the neural network parameters $\bm{\theta}$ gives:$$
\begin{aligned}
    \ell_{\mathrm{CR}}(\bm{\theta}) &= \mathbb{E}_{t, s, \bm c} \left[ \tfrac{1}{(s-t)^2} \mathbb{E}_{\bm{x}, \bm{\epsilon}} \left[ \Vert{} \bm\Delta \Vert{}^2_2 \mid \bm z_t, \bm c, t, s \right] \right],\\
    \bm\Delta  &= \bm f_{\bm{\theta}}(\bm z_t, t, s \mid \bm c) - \bm z_s.
\end{aligned}
$$
Since the factor $1/(s-t)^2$ is positive and independent of $\bm f_{\bm\theta}$, it does not affect the population minimizer for any fixed $t\neq s$. 
By standard properties of the $L_2$ loss, the optimal population minimizer $\bm f^\star$ is strictly given by the conditional expectation of the target $\bm z_s$ given the input state $\bm z_t$ and condition $\bm c$:$$
\bm f^*(\bm z_t, t, s \mid \bm c) = \mathbb{E}_{\bm{x}, \bm{\epsilon}} [\bm z_s \mid \bm z_t, \bm c].
$$
To analyze the behavior of this optimal minimizer for small time steps, we perform a Taylor expansion of  $\bm f^*(\bm z_t, t, s \mid \bm c)$  with respect to $s$ around $t$:
$$
\begin{aligned}
    \bm f^*(\bm z_t,t,s \mid \bm c)
    ={}& \bm f^*(\bm z_t,t,t\mid \bm c) \\
    &+ (s-t)
    \left.
    \frac{\partial}{\partial s}
    \bm f^*(\bm z_t,t,s\mid \bm c)
    \right|_{s=t} \\
    &+ \mathcal{O}\left(|s-t|^2\right)\\
    ={}&  \mathbb{E}_{\bm{x}, \bm{\epsilon}} [\bm z_t \mid \bm z_t,  \bm c] \\
    &+ (s-t)
    \left.
    \frac{\partial}{\partial s}
    \mathbb{E}_{\bm{x}, \bm{\epsilon}} [\bm z_s \mid \bm z_t,\bm c]
    \right|_{s=t} \\
    &+ \mathcal{O}\left(|s-t|^2\right)\\
    ={}& \bm z_t \\
    &+ (s-t) \ 
    \underbrace{
    \mathbb{E}_{\bm{x}, \bm{\epsilon}}
    \left[\dot{\bm z}_t \mid \bm z_t,\bm c\right]
    }_{\text{definition of velocity field}}\\
    &+ \mathcal{O}\left(|s-t|^2\right)\\
    ={}& \bm z_t
    + (s-t)\ \bm v(\bm z_t, t\mid \bm c) + \mathcal{O}\left(|s-t|^2\right) .\\
\end{aligned}
$$
On the other hand, the Taylor expansion of the exact flow map $\bm f(\bm z_t, t, s\mid \bm c)$ around $t$ is:
$$
\begin{aligned}
    \bm f(\bm z_t, t, s\mid \bm c) 
    = {}& \bm f(\bm z_t, t, t\mid \bm c) \\
    &+ (s-t) \left. \frac{\partial}{\partial s} \bm f(\bm z_t, t, s\mid \bm c) \right|_{s=t}\\
    &+ \mathcal{O}(|s-t|^2)\\
    = {}& \bm z_t + (s-t)\ \bm v(\bm z_t,t\mid \bm c)+ \mathcal{O}(|s-t|^2).
\end{aligned}
$$
Finally, we compare the optimal surrogate minimizer $\bm f^*(\bm z_t, t, s\mid \bm c)$ with the exact true flow map $\bm f(\bm z_t, t, s\mid \bm c)$ and obtain:
$$
\bm f^*(\bm z_t, t, s\mid \bm c) - \bm f(\bm z_t, t, s\mid \bm c) = \mathcal{O}(|s-t|^2)
$$
This demonstrates that the optimal population minimizer for the objective $\ell_{\mathrm{CR}}$ approximates the exact flow map with an error strictly bounded by $\mathcal{O}(|s-t|^2)$.
\end{proof}

\subsection{Proof of Proposition 2}
\begin{proof}
The velocity field model $\bm v_{\bm{\theta}}$ used in the CFM objective can be represented as the time derivative of the flow map model $\bm f_{\bm{\theta}}$ evaluated at $s=t$:
$$\bm v_{\bm{\theta}}(\bm z_t, t \mid \bm c) = \left. \frac{\partial}{\partial s} \bm f_{\bm{\theta}}(\bm z_t,t,s\mid \bm c) \right|_{s=t}  .
$$
Taking the Taylor expansion of $\bm f_{\bm{\theta}}$ with respect to $s$ around $t$, we obtain:
$$
\begin{aligned}
\bm f_{\bm{\theta}}(\bm z_t, t, s\mid \bm c) 
    ={}& \bm f_{\bm{\theta}}(\bm z_t,t,t\mid \bm c)\\
    &+ (s-t)
    \left.
    \frac{\partial}{\partial s}
    \bm f_{\bm{\theta}}(\bm z_t,t,s\mid \bm c)
    \right|_{s=t} \\
    &+ \mathcal{O}\left(|s-t|^2\right)\\
    ={}& \bm z_t + (s-t)\bm v_{\bm{\theta}}(\bm z_t, t\mid \bm c) + \mathcal{O}(|s-t|^2).
\end{aligned}
$$
Besides, we perform a Taylor expansion of the target state $\bm z_s$ around time $t$:
$$
\bm z_s = \bm z_t + (s-t)\  \dot{\bm z}_t + \mathcal{O}(|s-t|^2)\\
$$
Substituting the expanded forms of $\bm f_{\bm{\theta}}$ and $\bm z_s$ back into the difference term in the objective function $\ell_{\mathrm{CR}}$:
$$
\begin{aligned}
    \bm\Delta  ={}& \bm f_{\bm{\theta}}(\bm z_t, t, s \mid \bm c) - \bm z_s\\
    ={}& \left[\bm z_t + (s-t)\bm v_{\bm{\theta}}(\bm z_t, t\mid \bm c) + \mathcal{O}(|s-t|^2)\right]\\
    &-\left[\bm z_t + (s-t)\  \dot{\bm z}_t + \mathcal{O}(|s-t|^2)\right]\\
    ={}& (s-t)\left[\bm v_{\bm{\theta}}(\bm z_t, t\mid \bm c)- \dot{\bm z}_t\right] + \mathcal{O}(|s-t|^2)\ .
\end{aligned}
$$
Substituting the difference $\bm\Delta$ into the objective function $\ell_{\mathrm{CR}}$:
$$
\begin{aligned}
\ell_{\mathrm{CR}}(\bm{\theta})
    &= \mathbb{E}_{\bm{x},\bm{\epsilon},t,s}
    \Bigg[ \frac{1}{(s-t)^2} \big\|\bm\Delta \big \|_2^2 \Bigg]\\
    &=\mathbb{E}_{\bm{x}, \bm{\epsilon}, t, s} \left[ \left\| \bm v_{\bm{\theta}}(\bm z_t, t\mid \bm c) - \dot{\bm z}_t + \mathcal{O}(|s-t|) \right\|^2_2 \right]
\end{aligned}
$$
Taking the limit as $s \rightarrow t$, the $\mathcal{O}(|s-t|)$ residual term vanishes. The expectation over the target time $s$ drops out, resulting in:
$$
\lim_{s \rightarrow t} \ell_{\mathrm{CR}}(\bm{\theta}) = \mathbb{E}_{\bm{x}, \bm{\epsilon}, t} \left[ \left\| \bm v_{\bm{\theta}}(\bm z_t, t \mid \bm c) - \dot{\bm z}_t\right\|^2_2 \right] = \ell_{\mathrm{CFM}}(\bm{\theta})
$$
\end{proof}

\section{Experimental Configuration}

\subsection{Data Availability}
We utilize the preprocessed data published by STAEformer \cite{liu2023staeformer}, accessible via their GitHub repository: https://github.com/XDZhelheim/STAEformer/tree/main/data.

\subsection{Preprocessing}

We use Z-score normalization for data preprocessing. Specifically, we calculate the mean and standard deviation of the traffic states data from the training set, and then apply the same normalization to the training, validation, and test sets.

\subsection{Hyperparameters}

The hyperparameters for various datasets are listed in Table~\ref{tab:model_hyperparams}, and their search space is shown in Table~\ref{tab:searching_space}. 
We employed Optuna~\cite{optuna_2019} for hyperparameter optimization and chose the combination with the minimum Huber loss.

\begin{table}[htbp]
    \centering
    \small
    \begin{tabular}{l|cccc}
        \toprule
               & \multicolumn{4}{c}{PEMS} \\
        Parameter & $03$ & $04$ & $07$ & $08$ \\
        \midrule
        EMA decay         & \multicolumn{4}{c}{$0.9995$} \\
        Learning rate     & \multicolumn{4}{c}{$0.0002$} \\
         Early stop     & \multicolumn{4}{c}{$10$} \\
        Batch size        & \multicolumn{4}{c}{$8$} \\
        clip\_grad        & $0.5$ & $0$ & $0.5$ & $1$ \\
        \midrule
        Model dim ($D$)       & $128$  & $256$  & $256$  & $256$ \\
        Attention heads       & $8$    & $4$    & $8$    & $4$ \\
        Transformer FFN dim     & $512$ & $2048$ & $1024$ & $1024$ \\
        Blocks ($L$)     & $8$    & $5$    & $10$    & $6$ \\
        Dropout rate           & $0.3$  & $0.2$  & $0.1$ & $0.2$ \\
        \midrule
        Adaptive weight ($c$) & \multicolumn{4}{c}{$0.001$} \\
        Adaptive weight ($p$) & \multicolumn{4}{c}{$1.25$}  \\
        \midrule
        $t,s$ sampler         & \multicolumn{4}{c}{$\mathrm{LogNormal}(\mu,\sigma)$} \\
        sampling mean ($\mu$) & $-1.4$ & $-1.2$ & $-1.6$ & $-2.4$ \\
        sampling std ($\sigma$) & $1.8$ & $1.6$ & $1.6$ & $1.2$ \\
        \midrule
        Ratio & $0.5$ & $0.5$ & $0.25$ & $0.75$\\
        \bottomrule
    \end{tabular}
     \caption{Model Hyperparameters for Different Datasets.}
    \label{tab:model_hyperparams}
\end{table}

\begin{table}[htbp]
    \centering
    \renewcommand{\arraystretch}{1.2}
    \small
    \begin{tabular}{c|c}
        \toprule
        Parameters & Searching Space \\
        \midrule
       Adaptive weight ($p$)      & $[0.75 ,1.75]_{\Delta=0.25}$ \\
        sampling mean ($\mu$)   & $[-3 ,3]_{\Delta=0.2}$\\
        sampling std ($\sigma$)   & $[0.4, 2.2]_{\Delta=0.2}$ \\
        model dim ($D$)       & $\{128, 256\}$ \\
        Transformer FFN dim         & $\{256, 512, 1024, 2056\}$ \\
        Attention heads & $\{2, 4, 8\}$ \\
        Blocks ($L$)    & $[4 ,10]_{\Delta=1}$ \\
        Dropout rate    & $[0,0.3]_{\Delta=0.1}$ \\
        Ratio           & $[0,0.75]_{\Delta=0.25}$ \\
        \midrule
        EMA decay         & $0.9995$ \\
        Learning rate     & $0.0002$ \\
        Batch size        & $8$ \\
        Early stop         & $10$ \\
        Adaptive weight ($c$) & $0.001$\\
        \bottomrule
    \end{tabular}
    \caption{Hyperparameter Searching Space}
    \label{tab:searching_space}
\end{table}

\subsection{Random Seeds and Repeated Runs}
We use three independent training seeds, $\{2024, 2025, 2026\}$. The seed is set before model initialization and reset before constructing the data loaders. During evaluation, the sampling seed is fixed to 0. Tables~\ref{tab:performance_comparison_det} and~\ref{tab:performance_comparison_gen_mean} use three runs per trainable model--dataset pair.

\subsection{Computing Environment}
The models were trained and evaluated on an NVIDIA GH200 GPU using PyTorch 2.12.1 and Python 3.13.5, with CUDA 13.0 on an AlmaLinux 9.8 operating system.

\section{Evaluation Metrics}

We evaluate deterministic models using Mean Absolute Error (MAE), Root Mean Squared Error (RMSE), and Mean Absolute Percentage Error (MAPE). For generative models, we additionally use the Continuous Ranked Probability Score (CRPS) to evaluate the accuracy of the predictive distribution.

Let $\mathcal{I}$ denote the set of all scalar forecast entries in the test set, including all test windows, spatial nodes, forecast horizons, and feature channels. For $i\in\mathcal{I}$, let $y_i$ denote the target value from $x_f$, and let $\hat{y}_i^{(n)}$ denote the corresponding value in the $n$-th predictive trajectory. For a generative model with $N$ trajectories, the point prediction is defined as 
\begin{equation} \bar{y}_i = \frac{1}{N} \sum_{n=1}^{N}\hat{y}_i^{(n)}. \end{equation}

For a deterministic model, $\bar{y}_i$ denotes its direct output. We use $N\in\{1,10\}$ for generative evaluation. When $N=1$, $\bar{y}_i$ is a single sample generated by the model. The point prediction metrics are defined as \begin{align} \operatorname{MAE} &= \frac{1}{|\mathcal{I}_{\mathrm{nz}}|} \sum_{i\in\mathcal{I}_{\mathrm{nz}}} |\bar{y}_i-y_i|, \\ \operatorname{RMSE} &= \sqrt{ \frac{1}{|\mathcal{I}_{\mathrm{nz}}|} \sum_{i\in\mathcal{I}_{\mathrm{nz}}} (\bar{y}_i-y_i)^2}, \\ \operatorname{MAPE} &= \frac{100}{|\mathcal{I}_{\mathrm{nz}}|} \sum_{i\in\mathcal{I}_{\mathrm{nz}}} \frac{|\bar{y}_i-y_i|}{|y_i|}, \end{align} where $\mathcal{I}_{\mathrm{nz}}
=\{i\in\mathcal{I}:y_i\neq 0\}$. Zero-valued targets are excluded following the common practice~\cite{liang2022basicts}. All metrics are averaged jointly over the complete $T'=12$ forecast horizon. 

For probabilistic evaluation, let $F_i$ denote the predictive distribution at entry $i$. Let $Z_i$ and $Z_i'$ be independent draws from $F_i$. CRPS has the energy representation \begin{equation} \operatorname{CRPS}(F_i,y_i) = \mathbb{E}|Z_i-y_i| - \frac{1}{2}\mathbb{E}|Z_i-Z_i'|. \end{equation} 

Following \citet{gneiting2007strictly}, we use CRPS to evaluate each scalar predictive marginal. We estimate CRPS from $N=10$ independent predictive samples using the fair ensemble estimator of \citet{ferro2014fair}: 

\begin{table*}[!t]
\centering
\renewcommand{\arraystretch}{1.2}
\small
\setlength{\tabcolsep}{3pt}
\begin{adjustbox}{max width=\textwidth}
\begin{tabular}{c|ccc|ccc|ccc|ccc}
\toprule
Model
& \multicolumn{3}{c|}{PEMS03}
& \multicolumn{3}{c|}{PEMS04}
& \multicolumn{3}{c|}{PEMS07}
& \multicolumn{3}{c}{PEMS08} \\
\cmidrule(lr){2-4}
\cmidrule(lr){5-7}
\cmidrule(lr){8-10}
\cmidrule(lr){11-13}
& RMSE & MAE & MAPE
& RMSE & MAE & MAPE
& RMSE & MAE & MAPE
& RMSE & MAE & MAPE \\
\midrule

HI
& $49.892 \pm 0.000$ & $32.619 \pm 0.000$ & $30.596 \pm 0.000$
& $61.659 \pm 0.000$ & $42.353 \pm 0.000$ & $29.922 \pm 0.000$
& $71.340 \pm 0.000$ & $49.288 \pm 0.000$ & $22.753 \pm 0.000$
& $50.452 \pm 0.000$ & $34.661 \pm 0.000$ & $21.631 \pm 0.000$ \\

GWNet
& \underline{$25.080 \pm 0.094$}
& $\underline{14.545 \pm 0.019}$
& $14.857 \pm 0.153$
& $30.310 \pm 0.069$
& $18.948 \pm 0.061$
& $13.469 \pm 0.449$
& $33.517 \pm 0.087$
& $20.472 \pm 0.119$
& $8.683 \pm 0.117$
& $23.398 \pm 0.044$
& $14.586 \pm 0.007$
& $9.659 \pm 0.268$ \\

MTGNN
& $25.653 \pm 0.352$
& $14.953 \pm 0.061$
& $15.523 \pm 0.255$
& $30.940 \pm 0.020$
& $19.084 \pm 0.019$
& $13.407 \pm 0.197$
& $34.102 \pm 0.089$
& $20.958 \pm 0.071$
& $8.985 \pm 0.062$
& $24.281 \pm 0.094$
& $15.252 \pm 0.049$
& $10.001 \pm 0.157$ \\

STNorm
& $26.129 \pm 0.337$
& $15.275 \pm 0.055$
& \underline{$14.817 \pm 0.296$}
& $31.440 \pm 0.135$
& $19.056 \pm 0.078$
& $12.861 \pm 0.172$
& $34.488 \pm 0.299$
& $20.508 \pm 0.094$
& $8.639 \pm 0.033$
& $24.860 \pm 0.123$
& $15.406 \pm 0.078$
& $9.838 \pm 0.067$ \\

STID
& $27.606 \pm 0.746$
& $15.298 \pm 0.073$
& $16.166 \pm 0.178$
& \underline{$29.909 \pm 0.048$}
& $18.402 \pm 0.031$
& $12.573 \pm 0.045$
& $32.741 \pm 0.074$
& $19.645 \pm 0.031$
& $8.314 \pm 0.026$
& $23.350 \pm 0.068$
& $14.226 \pm 0.008$
& $9.356 \pm 0.054$ \\

STAEFormer
& $25.549 \pm 0.257$
& $15.010 \pm 0.058$
& $15.415 \pm 0.254$
& $30.199 \pm 0.272$
& \underline{$18.154 \pm 0.072$}
& $12.455 \pm 0.064$
& $32.891 \pm 0.181$
& $19.438 \pm 0.139$
& \underline{$8.028 \pm 0.024$}
& \underline{$23.225 \pm 0.049$}
& \underline{$13.511 \pm 0.058$}
& \underline{$8.858 \pm 0.016$} \\

HimNet
& $26.898 \pm 0.320$
& $15.182 \pm 0.125$
& $15.730 \pm 0.387$
& $30.077 \pm 0.120$
& $18.234 \pm 0.072$
& \underline{$12.076 \pm 0.111$}
& \underline{$32.620 \pm 0.034$}
& \underline{$19.238 \pm 0.014$}
& $8.276 \pm 0.180$
& $23.322 \pm 0.051$
& $13.592 \pm 0.016$
& $8.987 \pm 0.043$ \\

HyperD
& $\mathbf{24.205 \pm 0.104}$
& $14.940 \pm 0.054$
& $16.023 \pm 0.124$
& $30.018 \pm 0.082$
& $18.240 \pm 0.042$
& $12.510 \pm 0.041$
& $33.278 \pm 0.021$
& $19.434 \pm 0.010$
& $8.116 \pm 0.025$
& $23.318 \pm 0.012$
& $13.640 \pm 0.005$
& $9.016 \pm 0.007$ \\

\midrule

\textbf{CoReFlow}
& $25.321 \pm 0.156$
& $\mathbf{14.460 \pm 0.051}$
& $\mathbf{14.389 \pm 0.093}$
& $\mathbf{29.880 \pm 0.043}$
& $\mathbf{17.936 \pm 0.048}$
& $\mathbf{11.774 \pm 0.083}$
& $\mathbf{32.601 \pm 0.025}$
& $\mathbf{18.783 \pm 0.034}$
& $\mathbf{7.752 \pm 0.010}$
& $\mathbf{22.875 \pm 0.028}$
& $\mathbf{13.197 \pm 0.007}$
& $\mathbf{8.581 \pm 0.043}$ \\

\bottomrule
\end{tabular}
\end{adjustbox}
\caption{
Performance comparison with deterministic models.
CoReFlow is evaluated using one-sample generation ($N{=}1$).
All results are reported as mean $\pm$ standard deviation over three runs with different random seeds.
Best and second-best mean values are shown in bold and underlined, respectively.
}
\label{tab:performance_comparison_det}
\end{table*}

\begin{equation}
\begin{aligned}
\widehat{\operatorname{CRPS}}
={}& \frac{1}{|\mathcal{I}_{\mathrm{nz}}|}
\sum_{i\in\mathcal{I}_{\mathrm{nz}}}
\Bigg[
\frac{1}{N}\sum_{n=1}^{N}
\left|\hat{y}_i^{(n)}-y_i\right|
\\
&\quad
-\frac{1}{2N(N-1)}
\sum_{n=1}^{N}
\sum_{\substack{r=1\\ r\neq n}}^{N}
\left|\hat{y}_i^{(n)}-\hat{y}_i^{(r)}\right|
\Bigg].
\end{aligned}
\end{equation}

The denominator $N(N-1)$ excludes self-pairs and provides an unbiased estimate of the CRPS of the underlying predictive distribution. This correction avoids the finite-sample bias induced by treating the predictive samples as a finite empirical distribution. Lower values indicate better performance for all four metrics.

\section{Additional Experimental Results}
We report the mean and standard deviation over three runs with different random seeds. The complete results for deterministic and generative predictions are presented in Tables~\ref{tab:performance_comparison_det} and~\ref{tab:performance_comparison_gen_mean}, respectively.

\begin{table}[!t]
\centering
\small
\renewcommand{\arraystretch}{1.2}
\setlength{\tabcolsep}{4pt}
\begin{adjustbox}{max width=\columnwidth}
\begin{tabular}{c|cc|ccc|c}
\toprule
\multicolumn{3}{c}{Metric}
& DiffSTG & USTD & SpecSTG & \textbf{CoReFlow} \\
\midrule
    \multirow{7}{*}{\rotatebox{90}{PEMS07}}
    & \multirow{3}{*}{\rotatebox{90}{1 sample}}
    & RMSE
    & $55.926 \pm 6.590$
    & \underline{$40.747 \pm 1.351$}
    & $49.917 \pm 0.615$
    & $\mathbf{32.601 \pm 0.025}$ \\

    & & MAE
    & $39.047 \pm 5.587$
    & \underline{$26.602 \pm 1.137$}
    & $35.630 \pm 0.454$
    & $\mathbf{18.783 \pm 0.034}$ \\

    & & MAPE
    & $21.431 \pm 5.838$
    & \underline{$12.699 \pm 0.695$}
    & $19.172 \pm 0.061$
    & $\mathbf{7.752 \pm 0.010}$ \\

    \cmidrule(lr){2-7}

    & \multirow{4}{*}{\rotatebox{90}{10 samples}}
    & RMSE
    & $44.802 \pm 5.500$
    & \underline{$36.934 \pm 0.434$}
    & $39.458 \pm 0.860$
    & $\mathbf{32.562 \pm 0.025}$ \\

    & & MAE
    & $31.085 \pm 4.766$
    & \underline{$24.297 \pm 0.602$}
    & $26.298 \pm 0.693$
    & $\mathbf{18.731 \pm 0.036}$ \\

    & & MAPE
    & $18.430 \pm 5.809$
    & \underline{$12.060 \pm 0.625$}
    & $13.759 \pm 0.082$
    & $\mathbf{7.693 \pm 0.009}$ \\

    & & CRPS
    & $21.335 \pm 3.230$
    & $19.226 \pm 0.281$
    & \underline{$18.527 \pm 0.514$}
    & $\mathbf{18.201 \pm 0.038}$ \\

    \midrule

    \multirow{7}{*}{\rotatebox{90}{PEMS08}}
    & \multirow{3}{*}{\rotatebox{90}{1 sample}}
    & RMSE
    & $34.577 \pm 0.776$
    & \underline{$30.707 \pm 1.632$}
    & $34.045 \pm 1.237$
    & $\mathbf{22.875 \pm 0.028}$ \\

    & & MAE
    & $22.921 \pm 0.666$
    & \underline{$19.801 \pm 1.360$}
    & $24.370 \pm 0.912$
    & $\mathbf{13.197 \pm 0.007}$ \\

    & & MAPE
    & $14.561 \pm 0.254$
    & \underline{$13.801 \pm 0.964$}
    & $21.507 \pm 1.531$
    & $\mathbf{8.581 \pm 0.043}$ \\

    \cmidrule(lr){2-7}

    & \multirow{4}{*}{\rotatebox{90}{10 samples}}
    & RMSE
    & $27.334 \pm 0.404$
    & \underline{$26.892 \pm 1.124$}
    & $27.204 \pm 0.735$
    & $\mathbf{22.843 \pm 0.028}$ \\

    & & MAE
    & $17.921 \pm 0.380$
    & \underline{$17.456 \pm 1.017$}
    & $18.175 \pm 0.514$
    & $\mathbf{13.163 \pm 0.008}$ \\

    & & MAPE
    & \underline{$11.511 \pm 0.089$}
    & $12.780 \pm 0.837$
    & $16.106 \pm 1.349$
    & $\mathbf{8.546 \pm 0.040}$ \\

    & & CRPS
    & $14.016 \pm 0.227$
    & $13.161 \pm 0.650$
    & \underline{$12.817 \pm 0.323$}
    & $\mathbf{12.741 \pm 0.022}$ \\

    \bottomrule
\end{tabular}
\end{adjustbox}
\caption{
Performance comparison with generative models.
All results are reported as mean $\pm$ standard deviation over three runs with different random seeds.
Best and second-best mean values are shown in bold and underlined, respectively.
}
\label{tab:performance_comparison_gen_mean}

\end{table}

\section{Inference Efficiency}
\begin{figure}[!t]
    \centering
    \includegraphics[width=\columnwidth]{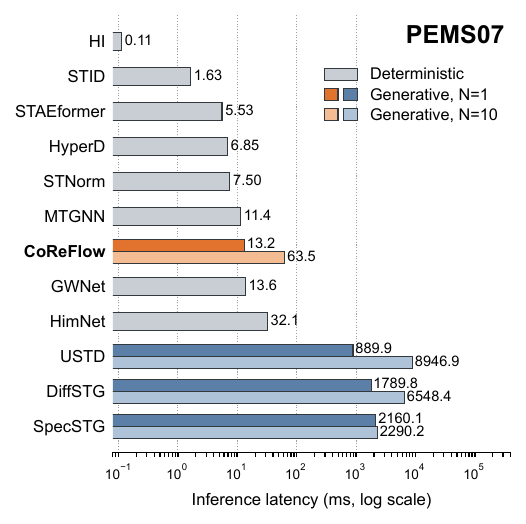}
    \caption{
    The full results of inference latency on PEMS07.}
    \label{fig:inference_latency07}
\end{figure}

\begin{figure}[!t]
    \centering
    \includegraphics[width=\columnwidth]{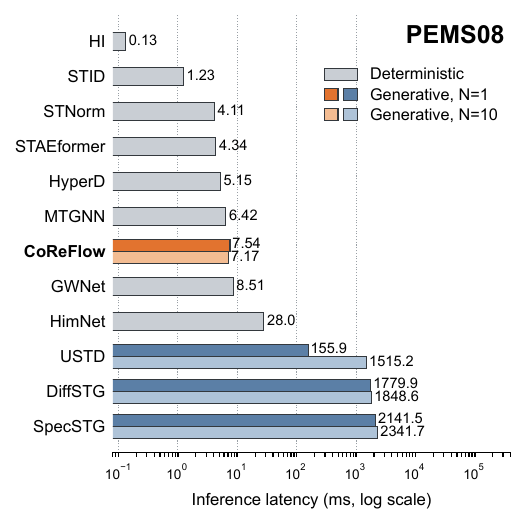}
    \caption{
    The full results of inference latency on PEMS08.}
    \label{fig:inference_latency08}
\end{figure}

We evaluate the inference latency of all models. For deterministic models, we measure the latency for producing a complete 12-step point forecast. For generative models, we measure the end-to-end latency for generating the complete 12-step forecast with $N\in\{1,10\}$ trajectories. We use a batch size of 1 and report the median over 200 runs after 50 warm-up runs on NVIDIA GH200 GPUs. Data loading, checkpoint loading, and metric computation are excluded. The complete results on PEMS07 and PEMS08 are shown in Fig.~\ref{fig:inference_latency07} and Fig.~\ref{fig:inference_latency08}, respectively.

\end{document}